\documentclass{article} 
\usepackage{tencent_ailab_tech_report}
\usepackage[colorlinks = true,
            linkcolor = blue,
            urlcolor  = blue,
            citecolor = blue,
            anchorcolor = blue]{hyperref}   
\usepackage{microtype}
\usepackage{hyperref}
\usepackage{url}
\usepackage{booktabs}
\usepackage{enumitem}
\usepackage{multicol}
\usepackage{multirow}
\usepackage{CJKutf8}
\usepackage{amsmath}
\usepackage{siunitx}
\usepackage{floatflt}
\usepackage{wrapfig}

\newcommand{\starcell}[1]{#1\rlap{${}^*$}}

\colmfinalcopy

\title{Scaling Synthetic Data Creation with 1,000,000,000 Personas}

\author{Tao Ge\thanks{Project lead (\tt{sggetao@gmail.com})}, ~Xin Chan, Xiaoyang Wang, Dian Yu, Haitao Mi, Dong Yu \\
Tencent AI Lab Seattle \\
\url{https://github.com/tencent-ailab/persona-hub}
}

\begin{document}

\maketitle

\begin{abstract} %

We propose a novel persona-driven data synthesis methodology that leverages various perspectives within a large language model (LLM) to create diverse synthetic data. To fully exploit this methodology at scale, we introduce Persona Hub -- a collection of 1 billion diverse personas automatically curated from web data. These 1 billion personas ($\sim$13\% of the world's total population), acting as distributed carriers of world knowledge, can tap into almost every perspective encapsulated within the LLM, thereby facilitating the creation of diverse synthetic data at scale for various scenarios. By showcasing Persona Hub's use cases in synthesizing high-quality mathematical and logical reasoning problems, instructions (i.e., user prompts), knowledge-rich texts, game NPCs and tools (functions) at scale, we demonstrate persona-driven data synthesis is versatile, scalable, flexible, and easy to use, potentially driving a paradigm shift in synthetic data creation and applications in practice, which may have a profound impact on LLM research and development.

\vspace{0.1cm}

\textit{\textcolor{red}{\textbf{DISCLAIMER}: \textup{Persona Hub} can facilitate synthetic data creation at a billion-scale to simulate diverse inputs (i.e., use cases) from a wide variety of real-world users. If this data is used as input to query a target LLM to obtain its outputs at scale, there is \textbf{a high risk} that the LLM's knowledge, intelligence and capabilities will be dumped and easily replicated, thereby challenging the leading position of the most powerful LLMs (e.g., our approach allows a 7B LLM to achieve 65\% on MATH, matching the performance of \textup{\texttt{gpt-4-turbo-preview}}). This tech report is \textbf{for research purposes only}. It is crucial to avoid misuse and ensure ethical and responsible application. We discuss its broad impact and potential concerns in detail in Section \ref{sec:concern}.}}
\end{abstract}

\vspace{-0.6cm}
\begin{figure}[!h]
    \centering
    \hspace*{0cm} \includegraphics[width=14cm]{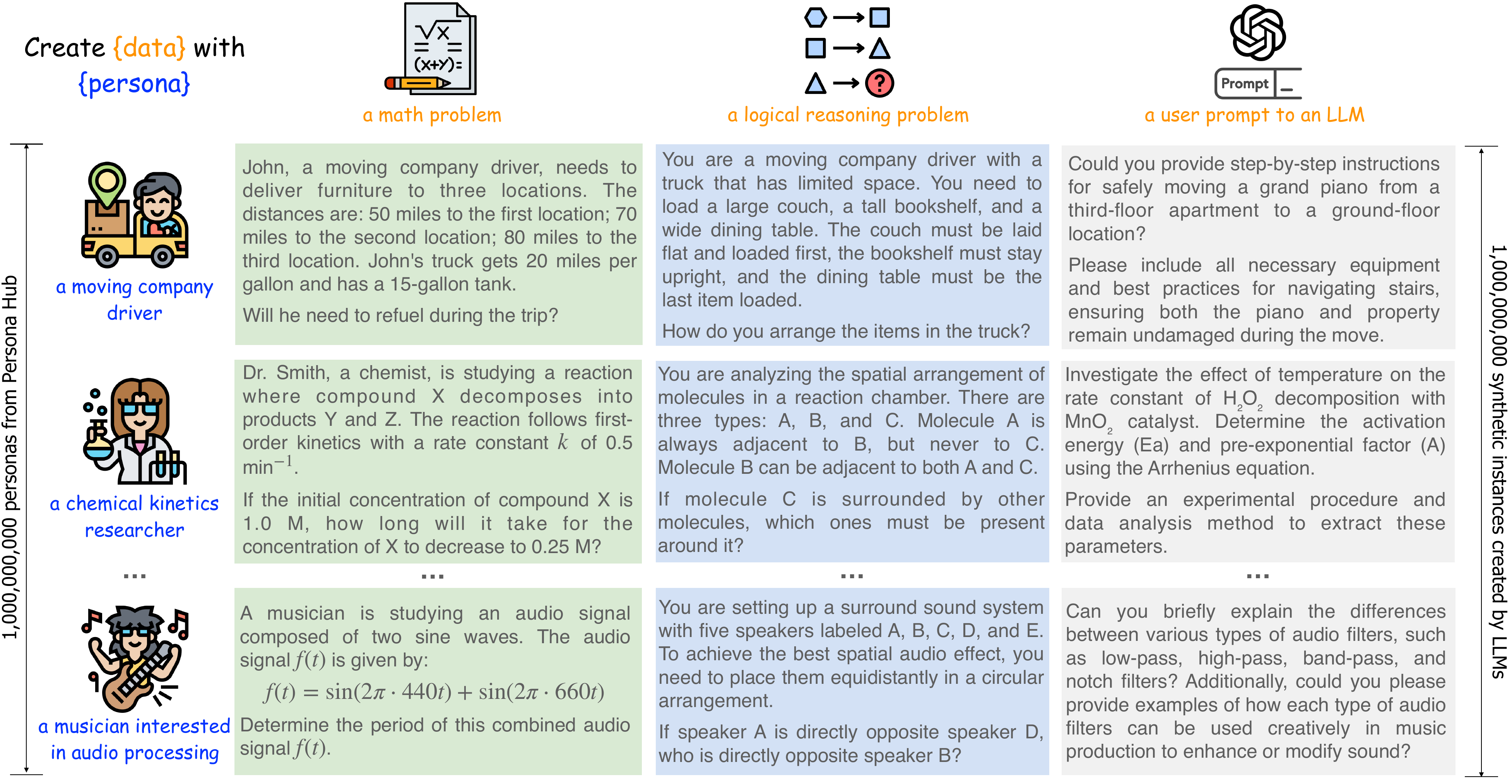}%
    \vspace{-0.2cm}
    \caption{Personas can work with a wide range of data synthesis prompts (e.g., ``create a math problem'') to guide an LLM to synthesize data with corresponding perspectives. The 1 billion personas in Persona Hub can facilitate various data synthesis scenarios at a billion scale.}
    \label{fig:overview}
\end{figure}

\section{Introduction}

As synthetic data~\citep{bauer2024comprehensive,liu2024best}, typically referring to data generated by models or algorithms rather than directly by humans, becomes increasingly valued~\citep{li2023textbooks} for training large language models (LLMs), there is a growing interest in data synthesis using LLMs: by simply specifying a data synthesis prompt, an LLM is expected to produce desirable synthetic data.

In practice, however, it is non-trivial to create synthetic data at scale: while we can easily scale up the quantity of synthetic data, it is difficult to ensure its diversity scales up as well. Without considering sampling\footnote{Sampling is orthogonal to this work. The diversity it introduces when solely used for data synthesis is usually limited.}, an LLM can only produce 1 instance given a data synthesis prompt. Therefore, to create diverse synthetic data at scale (e.g., 1 billion diverse math problems), a large number of diverse prompts are needed. 

Previous research tends to diversify the data synthesis prompt through the following two paradigms, but unfortunately, neither can practically achieve scalable synthetic data creation:


\begin{itemize}[left=5pt]
\item \textbf{Instance-driven}: This approach diversifies the data synthesis prompt by leveraging a seed corpus (i.e., creating new instances based on the instances in the seed corpus). Representative studies include \citet{wang2022self} and \citet{yu2023metamath}. However, under this paradigm, the diversity of the synthesized data mainly comes from the seed instances, making it difficult to truly extend beyond the seed corpus. Given the limited size of a seed corpus in most practical scenarios, it is challenging for this paradigm to scale up the creation of synthetic data.
\item \textbf{Key-point-driven}: This approach diversifies the data synthesis prompt with a curated comprehensive list of key points (or concepts) that can be a topic, a subject, or any knowledge we expect synthetic data to encompass. Representative studies include \citet{li2024synthetic} and \citet{huang2024key}. However, this methodology also faces difficulties in scaling synthetic data creation: it is practically prohibitive to curate a comprehensive list by enumerating all key points across different levels of granularity, unless limited to a narrow and specific domain (e.g., mathematics).
\end{itemize}

To practically achieve diverse synthetic data creation at scale, we propose a novel \textbf{persona-driven} data synthesis methodology. This is inspired by the observation that simply adding a persona to a data synthesis prompt can steer the LLM towards the corresponding perspective to create distinctive synthetic data, as shown in Figure \ref{fig:overview}. Since almost any LLM use case can be associated with a specific persona, we can create all-encompassing synthetic data at scale as long as we construct a comprehensive persona collection.

\begin{figure}[htbp]
    \centering
    \vspace{-0.3cm}
    \includegraphics[width=15cm]{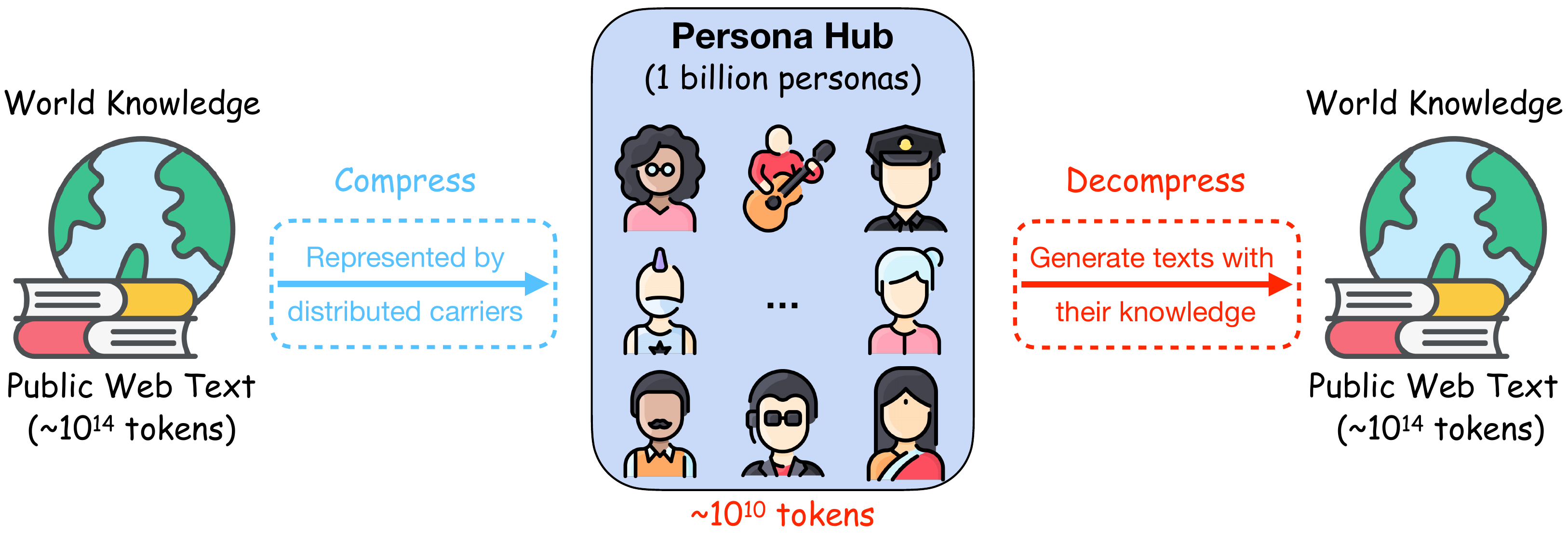}\vspace{-0.2cm}
    \caption{From a compression perspective~\citep{deletang2023language,ge2024incontext}, Persona Hub ($\sim 10^{10}$ tokens) can be seen as the compressed form of world knowledge (public web text for training LLMs, $\sim 10^{14}$ tokens) into distributed carriers. On the other hand, the public web text can be seen as the decompressed content created by these personas with their knowledge and experiences.}\label{fig:compression}
\end{figure}

Fortunately, personas are very easy to scale up. From massive web data, we automatically construct \textrm{Persona Hub} — a persona collection containing 1 billion diverse personas ($\sim$13\% of the world's total population). As Figure \ref{fig:compression} shows, these 1 billion personas can be regarded as distributed carriers of world knowledge, and each individual can be associated with their unique knowledge, experience, interest, personality and profession; thus, they can tap into almost every perspective encapsulated within the LLM to create diverse synthetic data at scale, without being limited by the size of a seed corpus. Moreover, in contrast to key points that typically work with specific data synthesis prompts, personas can be combined with almost any data synthesis prompt, benefiting from an LLM's strong roleplay ability~\citep{shanahan2023role,li2023steerability,choipicle,wang2024unleashing}, making them generally applicable to a variety of data synthesis scenarios.

We showcase \textrm{Persona Hub}'s use cases in large-scale creation of math and logical reasoning problems, instructions (i.e., user prompts), broad-coverage knowledge-rich texts, game NPCs, and tool (function) development. We demonstrate that persona-driven data synthesis is versatile, scalable, flexible, and easy to use, potentially driving a paradigm shift in synthetic data creation and applications in practice, which may have a profound impact on
LLM research and development.

To facilitate research in persona-driven data synthesis, we initially release \textbf{200,000 personas} from \textrm{Persona Hub} and following synthetic data samples we created with various personas, including:

\vspace{-0.15cm}
\begin{multicols}{2}
\begin{itemize}
    \item \textbf{50,000 math problems}
    \item \textbf{50,000 instructions}
    \item \textbf{10,000 game NPCs}
    \item \textbf{50,000 logical reasoning problems}
    \item \textbf{10,000 knowledge-rich texts}
    \item \textbf{5,000 tools (functions)}
\end{itemize}
\end{multicols}
\begin{itemize}
    \item \textbf{\textcolor{red}{[New in Feb 2025]} 370,000,000 expert personas with extraordinary expertise and skills}
\end{itemize}

\vspace{-0.25cm}

\textcolor{red}{\textbf{Note:} Our proposed methodology is applicable to almost any popular LLM\footnote{We mainly use publicly available LLMs such as GPT-4~\citep{achiam2023gpt}, Llama-3 and Qwen~\citep{qwen1.5, qwen2} in our experiments.}. The prompts shown in the figures throughout this paper are not exactly the prompt strings we used in our experiments; instead, they are simplified to fit the space and better illustrate the concepts. Interested readers can easily verify our methodology using the persona samples we have released. It is also worth noting that the main focus of this work is on creating new synthetic data, unlike much previous research that focuses on generating synthetic outputs for specific inputs (e.g., a math problem). Therefore, we use the terms ``create'' and ``synthesize'' interchangeably throughout the paper.}

\section{Persona Hub}

We propose two scalable approaches to derive diverse personas to construct \textrm{Persona Hub} from massive web data: \textit{Text-to-Persona} and \textit{Persona-to-Persona}.

\begin{figure}[htbp]
    \centering
    \includegraphics[width=15cm]{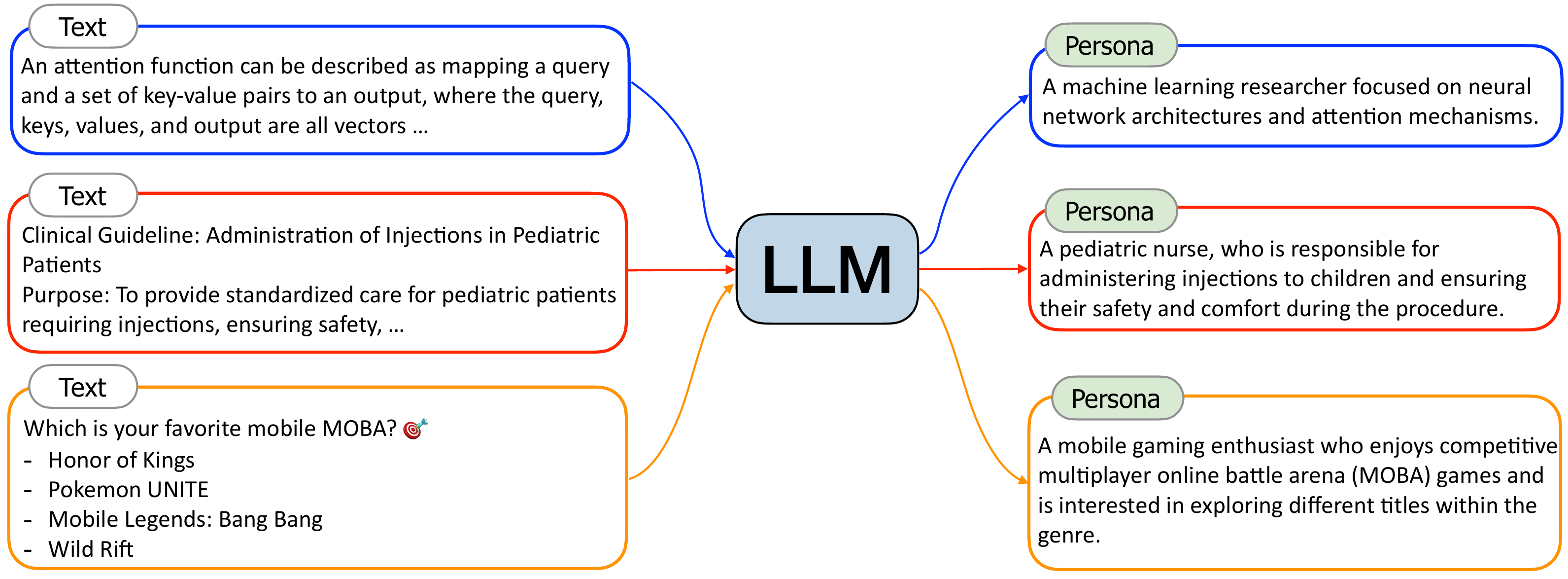}
    \caption{The \textit{Text-to-Persona} approach: it can use any text as input to obtain corresponding personas just by prompting the LLM ``Who is likely to [\texttt{read\textbar write\textbar like\textbar dislike\textbar\ldots}] the text?''}\vspace{-0.2cm}
    \label{fig:t2p1}
\end{figure}

\subsection{Text-to-Persona} \label{subsec:t2p}
A person with specific professional experiences and cultural backgrounds will have unique interests in reading and writing. Therefore, from a specific text, we can infer a specific persona who is likely to [\texttt{read\textbar write\textbar like\textbar dislike\textbar\ldots}] the text. Given that text data on the web is virtually unlimited and all-encompassing, we can obtain a wide-ranging collection of personas simply by prompting an LLM with these web texts, as shown in Figure \ref{fig:t2p1}.

\begin{figure}[t]
    \centering
    \includegraphics[width=15cm]{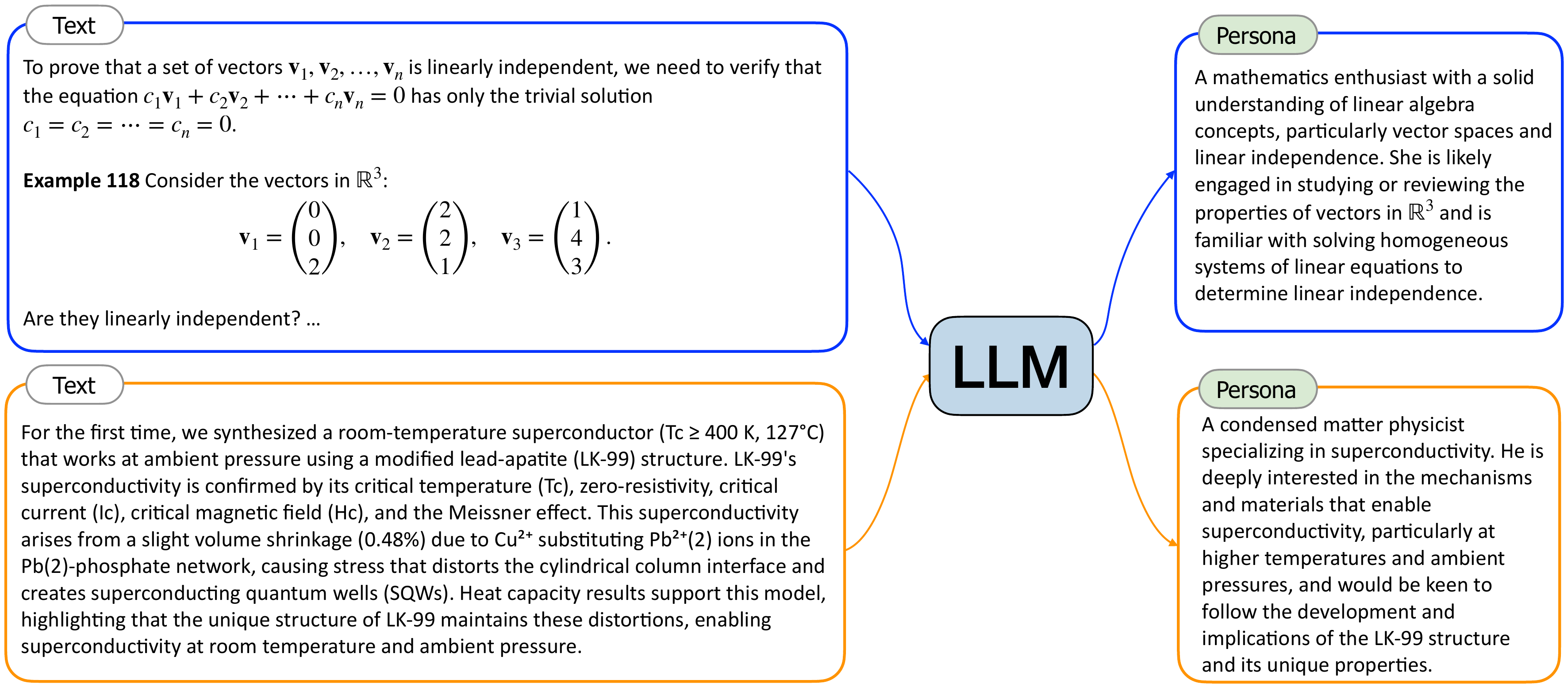}\vspace{-0.2cm}
    \caption{Persona descriptions will be fine-grained if input texts involve many detailed elements.}\vspace{-0.2cm}
    \label{fig:t2p2}
\end{figure}

There are many formats (e.g., plain text or structured text) to represent a persona, which can be controlled within the prompt. The granularity of an output persona description can also be adjusted through the prompt. For example, in the first case, a coarse-grained persona might be ``\textit{a computer scientist}'', whereas the fine-grained persona is ``\textit{a machine learning researcher focused on neural network architectures and attention mechanisms}''. In our practice, we ask the LLM (in the prompt) to output persona descriptions as specifically as possible. Besides specifying the granularity of persona descriptions in the prompt, input texts can also influence the granularity of persona descriptions. As shown in Figure \ref{fig:t2p2}, if an input text (e.g., from a mathematical textbook or an academic paper about superconductivity) contains many detailed elements, the resulting persona description will also be specific and fine-grained. Therefore, by applying the \textit{Text-to-Persona} approach to massive web text data, we can obtain billions (or even trillions) of diverse personas, encompassing a wide range of aspects across different granularities.

\subsection{Persona-to-Persona} \label{subsec:p2p}
As discussed above, \textit{Text-to-Persona} is a highly scalable method that can synthesize personas covering almost every aspect. However, it may still miss some personas that have low visibility on the web and thus are less likely to obtain via \textit{Text-to-Persona}, such as \textit{a child}, \textit{a beggar}, or \textit{a behind-the-scenes crew member of a movie}. To supplement the personas that \textit{Text-to-Persona} might hardly reach, we propose \textit{Persona-to-Persona}, which derives personas with interpersonal relationships from those obtained through \textit{Text-to-Persona}.

\begin{figure}[t]
    \centering
    \includegraphics[width=13cm]{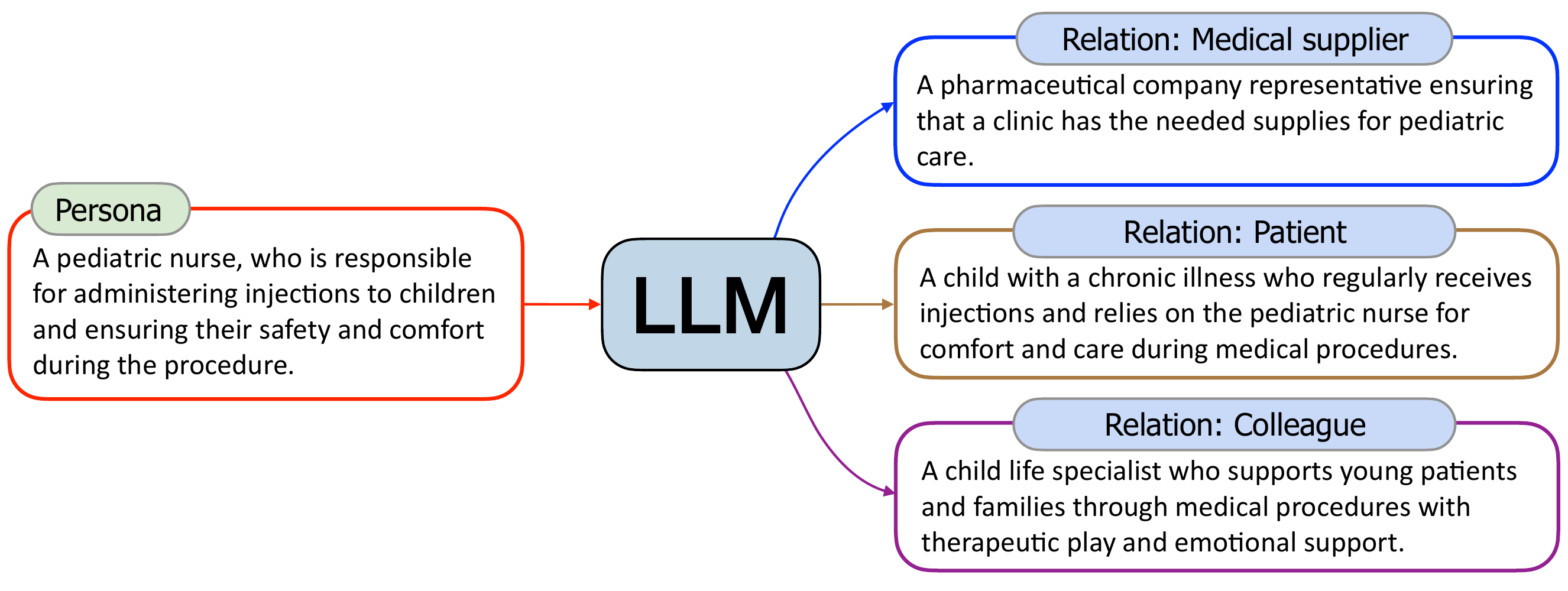}\vspace{-0.2cm}
    \caption{\textit{Persona-to-Persona} obtains diverse personas via interpersonal relationships, which can be easily achieved by prompting the LLM ``Who is in close relationship with the given persona?''}\vspace{-0.2cm}
    \label{fig:p2p}
\end{figure}

As shown in Figure \ref{fig:p2p}, the persona about ``\textit{a child}'' can be derived from the persona of a nurse at a children's hospital (patient-caregiver relationship). Similarly, ``\textit{a beggar}'' can be derived from the persona of a shelter worker (assistance relationship), and ``\textit{a behind-the-scenes movie crew member}'' can be derived from the persona of the movie's lead actor (co-worker relationship). According to the six degrees of separation theory~\citep{travers1977experimental}, we perform six iterations of persona relationship expansion for each persona obtained through \textit{Text-to-Persona}, thereby enriching our persona collection even further.

\subsection{Deduplication}
We first run \textit{Text-to-Persona} on the RedPajama v2 dataset~\citep{together2023redpajama} and then perform \textit{Persona-to-Persona}, as described in Sections \ref{subsec:t2p} and \ref{subsec:p2p}. After obtaining billions of personas, it is inevitable that some of the personas will be identical or extremely similar. To ensure the diversity of \textrm{Persona Hub}, we deduplicate these personas in two ways:

\paragraph{MinHash-based Deduplication}
We use MinHash~\citep{broder1997resemblance} to deduplicate based on the n-gram features of persona descriptions. Since persona descriptions are usually just 1-2 sentences, much shorter than a document, we simply used 1-gram and a signature size of 128 for MinHash deduplication. We deduplicate at the similarity threshold of 0.9.

\paragraph{Embedding-based Deduplication} 
After deduplication based on surface forms (i.e., MinHash with n-gram features), we also adopt embedding-based deduplication. We use a text embedding model (e.g., the \texttt{text-embedding-3-small} model from OpenAI) to compute an embedding for each persona, and then filter out personas with a cosine semantic similarity greater than 0.9.

Note that although we select 0.9 as the threshold here, we can flexibly adjust it according to specific needs for further deduplication. For instance, when the requirement for the number of instances is not high (e.g., only needing 1 million instances) but the demand for diversity is high, we can further apply a stricter deduplication standard (e.g., discarding personas with a similarity greater than 0.5).

After deduplication and using simple heuristic methods to filter out low-quality persona descriptions, we have harvested a total of 1,015,863,523 personas, finally forming our \textrm{Persona Hub}.

\section{Persona-driven Synthetic Data Creation}\label{sec:methodology}

Our proposed persona-driven data synthesis approach is straightforward and effective, which involves integrating a persona into the appropriate position in a data synthesis prompt. Simple as it appears, it can significantly influence the LLM to adopt the persona's perspective to create synthetic data. Driven by the 1 billion personas in \textrm{Persona Hub}, this approach can easily create diverse synthetic data at a billion scale.

\begin{figure}[t]
    \centering
    \includegraphics[width=14cm]{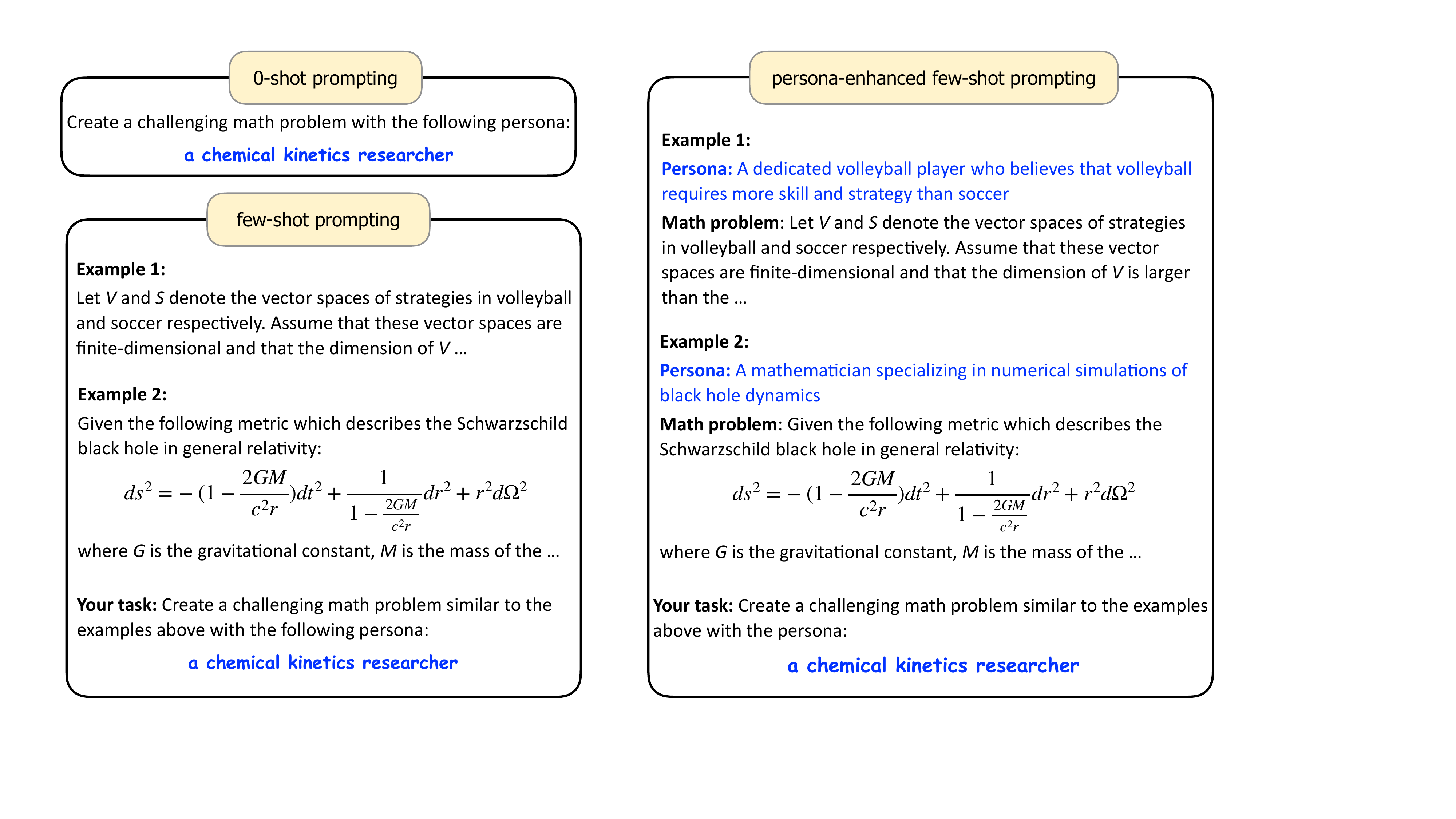}\vspace{-0.2cm}
    \caption{0-shot, few-shot and persona-enhanced few-shot prompting methods.}\vspace{-0.2cm}
    \label{fig:prompting}
\end{figure}

Just as we can use either zero-shot or few-shot methods to prompt an LLM, the persona-driven methodology is also flexible and compatible with various forms of prompts to create synthetic data. As shown in Figure \ref{fig:prompting}, we propose three persona-driven data synthesis prompting methods:

\begin{itemize}[left=5pt]
\item \textbf{Zero-shot prompting} does not leverage any existing examples (i.e., demonstrations), thereby fully exploiting the model's creativity without being constrained by specific examples.
\item \textbf{Few-shot prompting} can better ensure that the synthesized data meets the requirements by providing some demonstrations.
\item \textbf{Persona-enhanced few-shot prompting} is more effective in enhancing the LLM's persona-driven data synthesis capabilities. However, its drawback is that it requires deriving the corresponding persona for each demonstration in the few-shot prompt beforehand.
\end{itemize}

\section{Use Cases}\label{sec:usecases}

We demonstrate the use cases of \textrm{Persona Hub} in various data synthesis scenarios, including the large-scale creation of math and logical reasoning problems, instructions (i.e., user prompts), knowledge-rich texts, game NPCs, and tool (function) development.

As mentioned earlier, the persona-driven approach is general and versatile, making it easily adaptable to different data synthesis scenarios simply by adjusting the data synthesis prompt. Therefore, we will provide a detailed technical discussion only for math problem synthesis (Section \ref{subsec:math}) and skip the detailed discussion for other use cases.

\subsection{Math Problems}\label{subsec:math}

\subsubsection{Demonstrations} \label{subsubsec:mathdemo}

\begin{figure}[t]
    \centering
    \includegraphics[width=14cm]{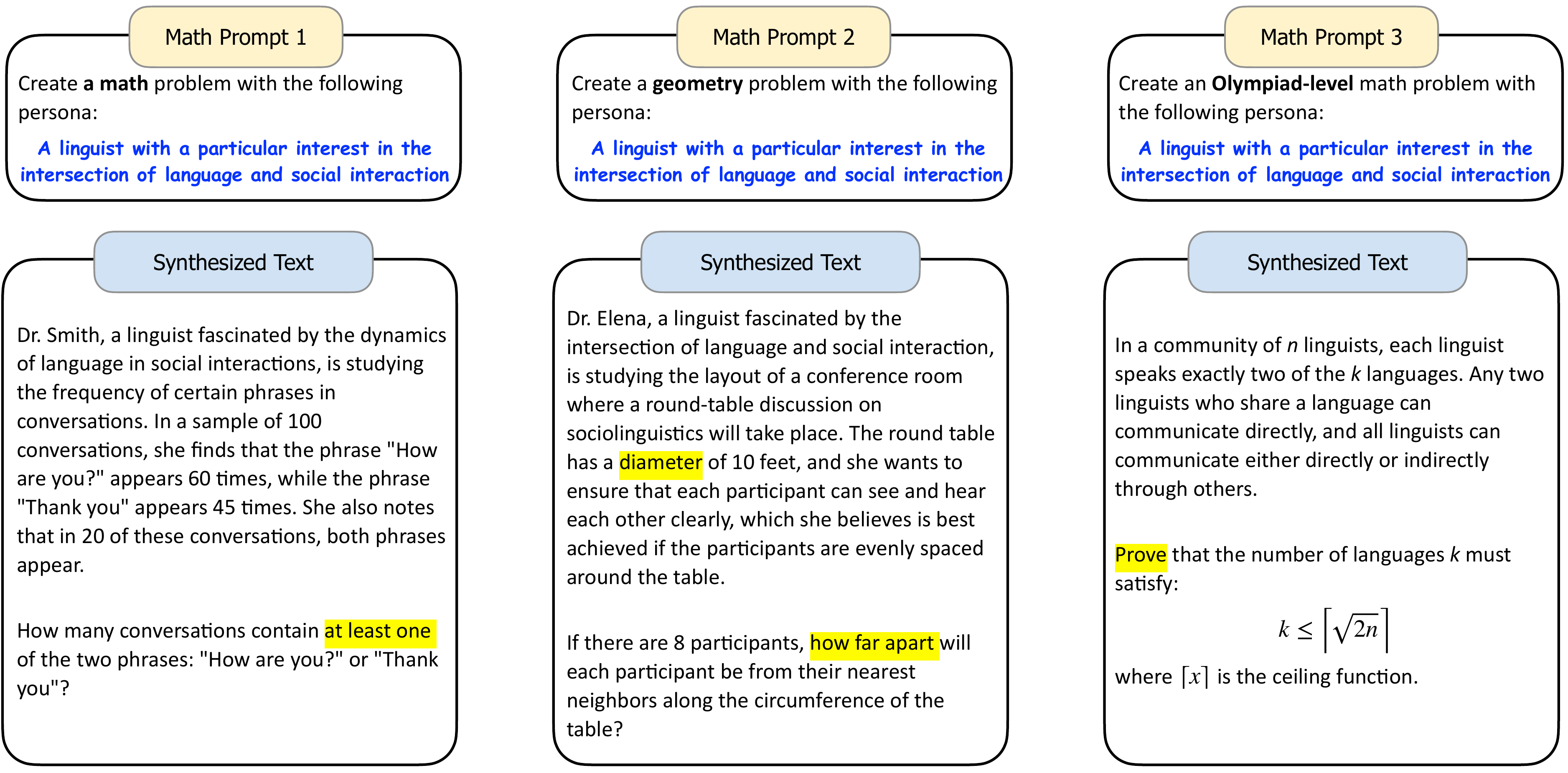}\vspace{-0.3cm}
    \caption{A linguist persona with different math problem creation prompts that specify the focus (e.g., geometry) or the difficulty (e.g., Olympiad-level)}
    \label{fig:flexibility}
\end{figure}

As the initial example (Figure \ref{fig:overview}) shows, when prompting an LLM to create a math problem, adding a persona leads the LLM to create math problems related to that persona. The example in Figure \ref{fig:flexibility}(left) further confirms this: when presented with a linguist persona, the LLM will create a math problem in the context of computational linguistics. Moreover, adding a persona does not hinder the flexibility of the prompt -- we can still easily specify the focus (Figure \ref{fig:flexibility}(middle)) or difficulty (Figure \ref{fig:flexibility}(right)) of our desired math problem in the prompt.

\begin{figure}[htbp]
    \centering
    \includegraphics[width=14cm]{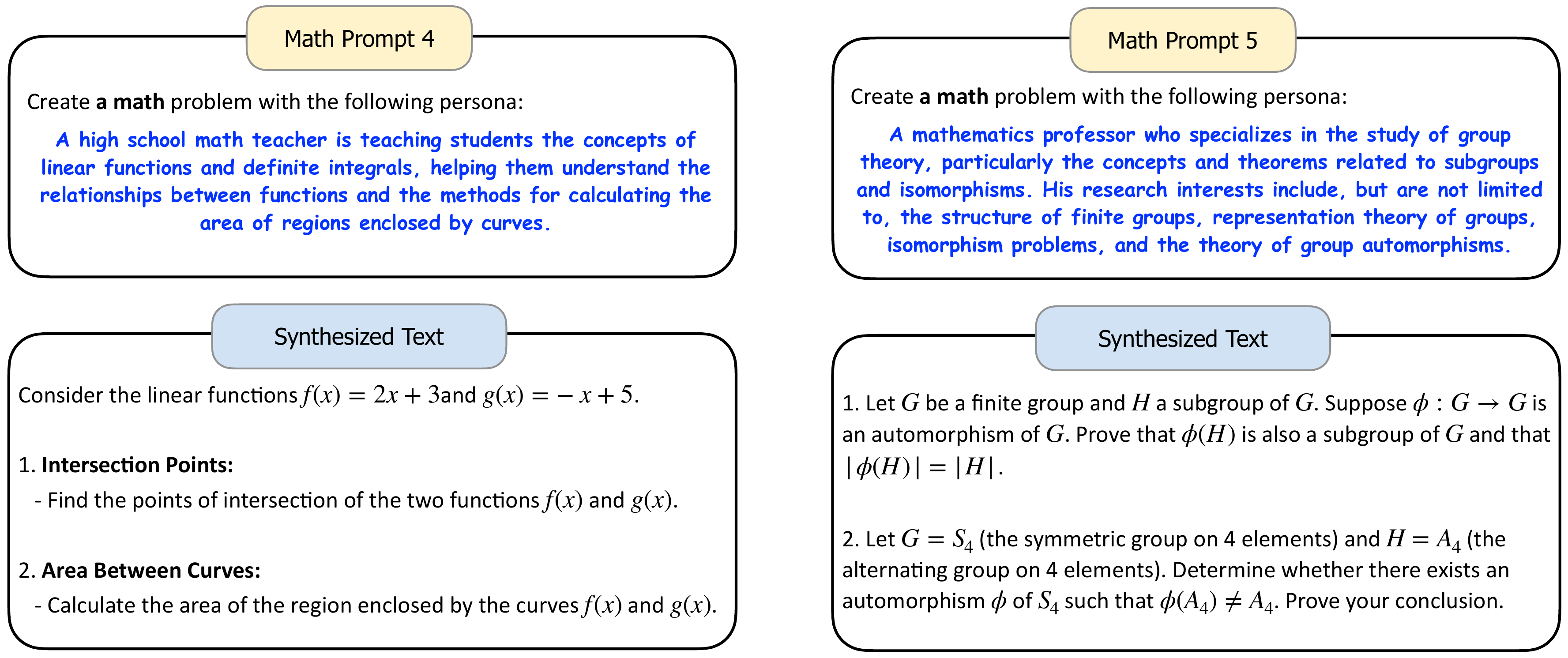}
    \caption{Examples of math problems created with personas of professionals related to the field of mathematics. They tend to be more challenging than those created with general personas because they usually require a deeper and more fine-grained understanding of advanced mathematical knowledge and skills.}
    \label{fig:mathexpert}
\end{figure}

The examples in Figure \ref{fig:flexibility} demonstrate the use of general personas to create math problems. We can certainly employ professionals related to mathematics to create math problems as well. As shown in Figure \ref{fig:mathexpert}, personas of math professionals\footnote{We can easily harvest a large number of such personas when running \textit{Text-to-Persona} (Section \ref{subsec:t2p}) on public web texts, particularly when processing texts in the field of mathematics.} often mention more advanced and granular mathematics knowledge and skills (as discussed earlier in Section \ref{subsec:t2p} and Figure \ref{fig:t2p2}), which in turn allows the created math problems to cover these mathematical concepts, making them more challenging.

\subsubsection{Evaluation} \label{subsubsec:matheval}

\paragraph{Data} We select 1.09 million personas from \textrm{Persona Hub} and employ the 0-shot prompting method using GPT-4 to create math problems with these personas, which does not leverage any instances from benchmarks like MATH~\citep{hendrycks2021measuring} during the creation of math problems. This approach allowed us to synthesize 1.09M math problems. Since this work focuses on creating new synthetic data rather than synthesizing solutions, we simply used \texttt{gpt-4o} (assistant\footnote{Assistant system message in OpenAI API doc: ``You are a helpful assistant.''}) to generate solutions to the created problems. Among these 1.09M math problems, we randomly hold out 20k as a synthetic test set to facilitate evaluation. The remaining 1.07M problems are used for training. We use the following two test sets for evaluation:

\begin{itemize}[left=5pt]
    \item \textbf{Synthetic Test Set} (In-distribution): Since the set of the held-out 20K problems is produced in the same way as the 1.07M training instances, it can be considered an in-distribution test set. To ensure the accuracy of the answers in this test set for increasing the reliability of the evaluation, we additionally generate solutions using \texttt{gpt-4o} (PoT\footnote{Program of thought prompting~\citep{chen2022program}}) and \texttt{gpt-4-turbo} (assistant) in addition to the solution generated by \texttt{gpt-4o} (assistant). We retain only the test instances where at least two solutions are consistent. The remaining test set consists of 11.6K test instances.
    
    \item \textbf{MATH} (Out-of-distribution): The most widely recognized benchmark for testing the mathematical reasoning ability of LLMs. Its test set contains 5,000 competitive-level math problems with reference answers. Since we do not use any instances from the MATH dataset for data synthesis or training, we regard the MATH test set as an out-of-distribution test set.
\end{itemize}

\begin{table}[h!]
\centering
\small
\begin{tabular}{lcc}
\toprule
\textbf{Model} & \textbf{Model Size} & \textbf{Accuracy (\%)} \\
\midrule
\multicolumn{3}{c}{\textbf{Open-sourced LLMs}} \\
\midrule
DeepSeek LLM 67B Chat~\citep{bi2024deepseek} & 67B & 53.2 \\
Phi-3-Mini-4K-Instruct \citep{abdin2024phi} & 3.8B & 68.3 \\
Yi-1.5-34B-Chat \citep{young2024yi} & 34B & 70.4 \\
Qwen1.5-72B-Chat \citep{qwen1.5} & 72B & 60.7 \\
Qwen1.5-110B-Chat \citep{qwen1.5} & 110B & 73.0 \\
Qwen2-7B-Instruct \citep{qwen2} & 7B & 72.1 \\
Qwen2-72B-Instruct \citep{qwen2} & 72B & 77.2 \\
Llama-3-8B-Instruct & 8B & 39.8 \\
Llama-3-70B-Instruct & 70B & 63.5 \\
\midrule
\multicolumn{3}{c}{\textbf{GPT-4}} \\
\midrule
\texttt{gpt-4-turbo-2024-04-09} & ? & 88.1 \\
\texttt{gpt-4o-2024-05-13} & ? & 91.2 \\
\midrule
\multicolumn{3}{c}{\textbf{This work}} \\
\midrule
Qwen2-7B (fine-tuned \textit{w/} the 1.07M synthesized instances) & \bf 7B & \bf 79.4 \\
\bottomrule
\end{tabular}\vspace{-0.2cm}
\caption{In-distribution evaluation results on the 11.6K synthetic test instances.}
\label{tab:id_result}
\end{table}

\paragraph{Equality Checking} We follow the same evaluation protocol as OpenAI\footnote{\url{https://github.com/openai/simple-evals}} to check answer equality on the MATH benchmark. For the synthetic test set, we use a similar method, except we use Llama-3-70B-Instruct instead of \texttt{gpt-4-turbo-preview} as the equality checker.

We simply fine-tune the latest open-sourced 7B LLM -- Qwen2-7B~\citep{qwen2} with our synthesized 1.07 million math problems and evaluate its greedy decoding outputs on the above two test sets.

\begin{table}[t!]
\centering
\small
\begin{tabular}{l c S[table-format=2.1] }
\toprule
\textbf{Model} & \textbf{Model Size} & \textbf{Accuracy (\%)} \\
\midrule
\multicolumn{3}{c}{\textbf{State-of-the-art LLMs}} \\
\midrule
\texttt{gpt-4o-2024-05-13} & ? & \textbf{76.6} \\
\texttt{gpt-4-turbo-2024-04-09} & ? & 73.4 \\
\texttt{gpt-4-turbo-0125-preview} & ? & 64.5 \\
\texttt{gpt-4-turbo-1106-preview} & ? & 64.3 \\
\texttt{gpt-4} & ? & \starcell{52.6} \\
Claude 3.5 Sonnet & ? & \starcell{71.1} \\
Claude 3 Opus & ? & 63.8 \\
Gemini Pro 1.5 (May 2024) & ? & \starcell{67.7} \\
Gemini Ultra & ? & \starcell{53.2} \\
DeepSeek-Coder-V2-Instruct \citep{zhu2024deepseek} & 236B/21B & \starcell{75.7} \\
Llama-3-70B-Instruct & 70B & 52.8 \\
Qwen2-72B-Instruct & 72B & \starcell{59.7} \\
Qwen2-7B-Instruct & 7B & \starcell{49.6} \\
\midrule
\multicolumn{3}{c}{\textbf{This work}} \\
\midrule
Qwen2-7B (fine-tuned \textit{w/} the 1.07M synthesized instances) & \bf 7B & \textbf{64.9} \\
\bottomrule
\end{tabular}\vspace{-0.1cm}
\caption{Out-of-distribution evaluation on MATH. Results marked with an asterisk (*) may not use the OpenAI's evaluation method. The model fine-tuned with our synthesized 1.07M math problems achieves 64.9\% on MATH, matching the performance of \texttt{gpt-4-turbo-preview} at only a 7B scale.}\label{tab:ood_result}\vspace{-0.6cm}
\end{table}

Table \ref{tab:id_result} presents the in-distribution (ID) evaluation results on the 11.6K synthetic test instances. Among the tested open-source LLMs, Qwen2-72B-Instruct achieves the best result, and the ranking of the other models is generally consistent with their reported performance on other mathematical benchmarks. Our model, with the help of the 1.07M synthetic math problems, achieves nearly 80\% accuracy, surpassing all the open-source LLMs. However, considering that the answers in the synthetic test are not absolutely reliable and that our model might be the only one using ID training data, this ID evaluation results should be taken as a reference only. 


We present the evaluation results on MATH in Table \ref{tab:ood_result}. The 7B model fine-tuned with the synthetic training data achieved an impressive 64.9\% accuracy on MATH simply using greedy decoding, outperformed only by \texttt{gpt-4o}, \texttt{gpt-4-turbo-2024-04-09}, Claude 3.5 Sonnet, Gemini Pro 1.5 (May 2024) and DeepSeek-Coder-V2-Instruct. 

\begin{wrapfigure}{r}{0.52\textwidth}
    \centering
    \includegraphics[width=0.52\textwidth]{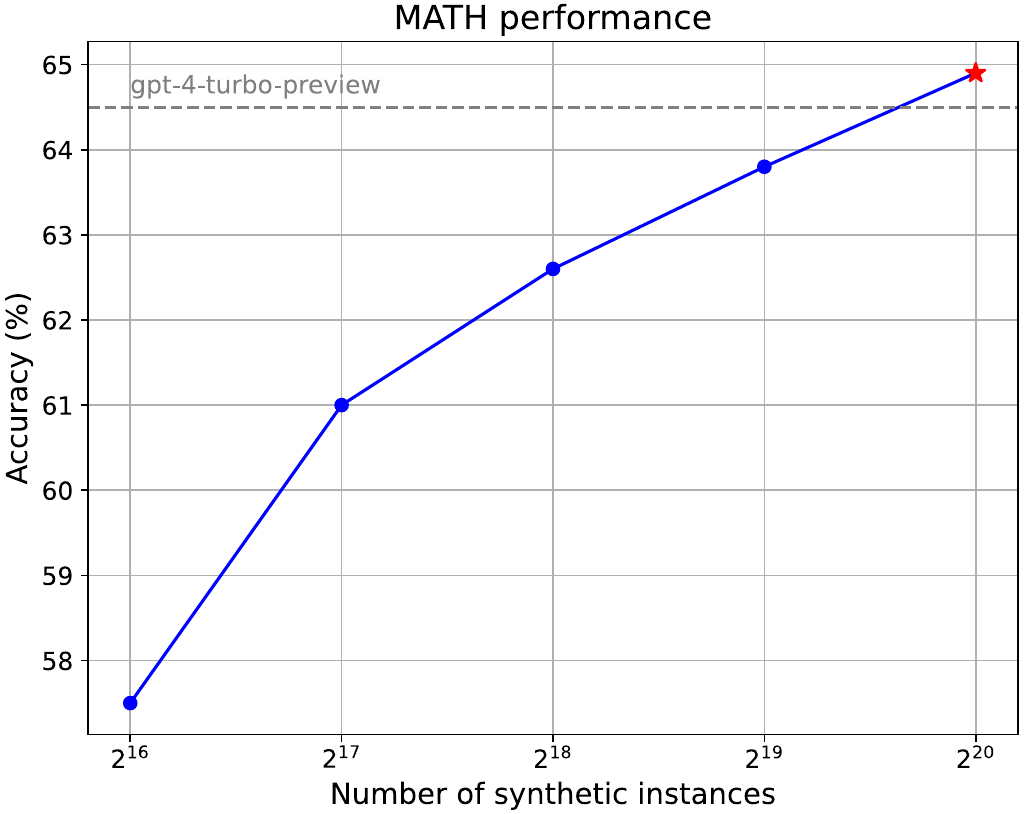}
    \caption{Accuracy on MATH with scaling the synthetic instances used for training Qwen2-7B}\vspace{-0.1cm}
    \label{fig:scaling}
\end{wrapfigure}

Figure \ref{fig:scaling} presents the performance of the model on MATH when trained with synthetic math problems at different scales. Its performance trend generally aligns with the scaling law~\citep{kaplan2020scaling}. Unlike previous research \citep{yu2023metamath,wang2023mathcoder,li2024common} that performs scaling on in-distribution data (e.g., heavily relying on MATH train data to augment in-distribution data), we did not use any instances from MATH during data synthesis or training. Therefore, in this out-of-distribution (OOD) evaluation setting, achieving performance on MATH that surpasses \texttt{gpt-4-turbo-preview (1106/0125)} is indeed impressive and promising for a 7B model.

\begin{figure}[htbp]
    \centering
    \includegraphics[width=15.2cm]{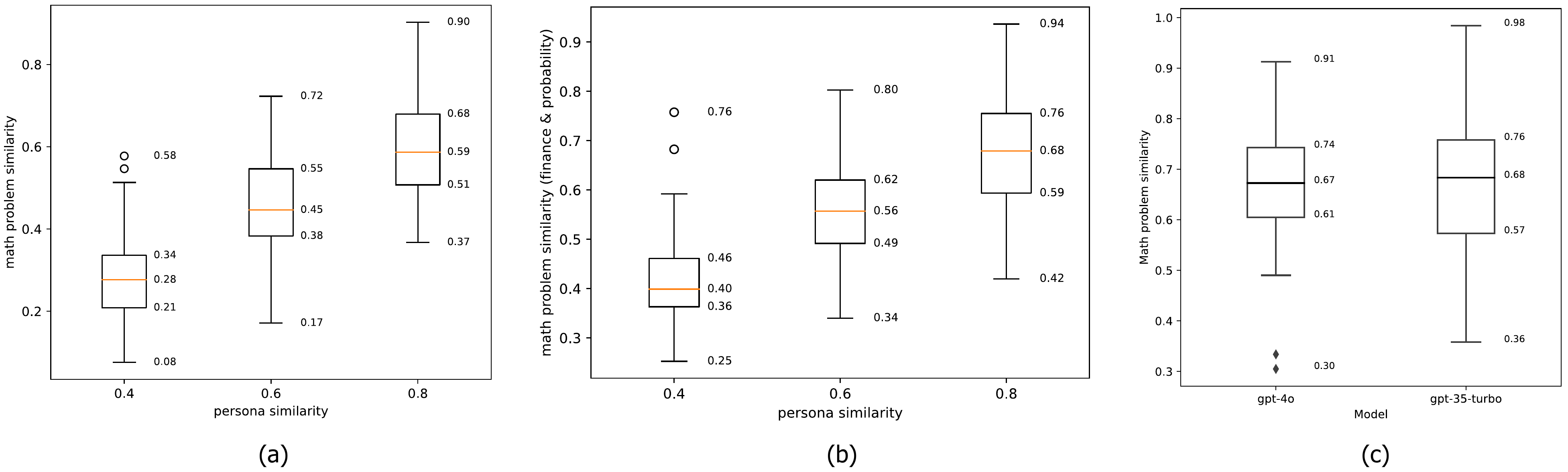}\vspace{-0.3cm}
    \caption{Similarities of math problems created by personas with different similarities: \textbf{(a)} Similarity of math problems when no specific focus is given; \textbf{(b)} Similarity of math problems when the prompt specifies they must be related to finance and probability; \textbf{(c)} Similarity of math problems synthesized by \texttt{gpt-4o} and \texttt{gpt-35-turbo} with persona similarity of 0.9.}\vspace{-0.2cm}
    \label{fig:sim}
\end{figure}

We examine the quality of our synthesized math problems: we sample 200 challenging problems (involving high school and university-level math knowledge points in China), and have two math experts evaluate their validity. Only 7 out of 200 problems are marked as invalid (e.g., due to insufficient or conflicting conditions), yielding a reliable validity rate of 96.5\%.

Moreover, we specifically examine the impact of differences in personas within the prompts on the synthesized math problems. We first sample 100 pairs of personas with semantic similarities\footnote{We use OpenAI's \texttt{text-embedding-3-small} (dim=512) to obtain the semantic representation and compute cosine similarity in this experiment. Here, a persona similarity of 0.4 means that a pair of personas has a semantic similarity within the range of 0.39 to 0.41.} of 0.4, 0.6, and 0.8, respectively. For each pair of personas, we use them to create a pair of math problems using greedy decoding (i.e., temperature=0). Then, we compute the semantic similarity of these math problem pairs and show the results in Figure \ref{fig:sim}.

We can clearly observe that the semantic similarity between synthesized math problems tends to be correlated with but lower than the similarity between their corresponding personas. When we add more specific constraints to the prompts (e.g., math problems about finance and probability), the similarity between the synthesized math problems tends to become higher (Figure \ref{fig:sim}(b)). In Figure \ref{fig:sim}(c), we also test the similarity of math problems created by \texttt{gpt-4o} and \texttt{gpt-35-turbo} using highly similar personas (similarity=0.9). The results indicate that the semantic similarity of the math problems created by \texttt{gpt-4o} and \texttt{gpt-35-turbo} seems not significantly different: most synthesized math problems' similarity falls within the range of 0.6 to 0.75, which is much lower than the similarity of the personas (0.9). Given these observations, we believe that using the personas in Persona Hub can ensure the diversity of synthesized data -- even at a billion scale.

\subsection{Logical Reasoning Problems}

\begin{figure}[t]
    \centering
    \includegraphics[width=13cm]{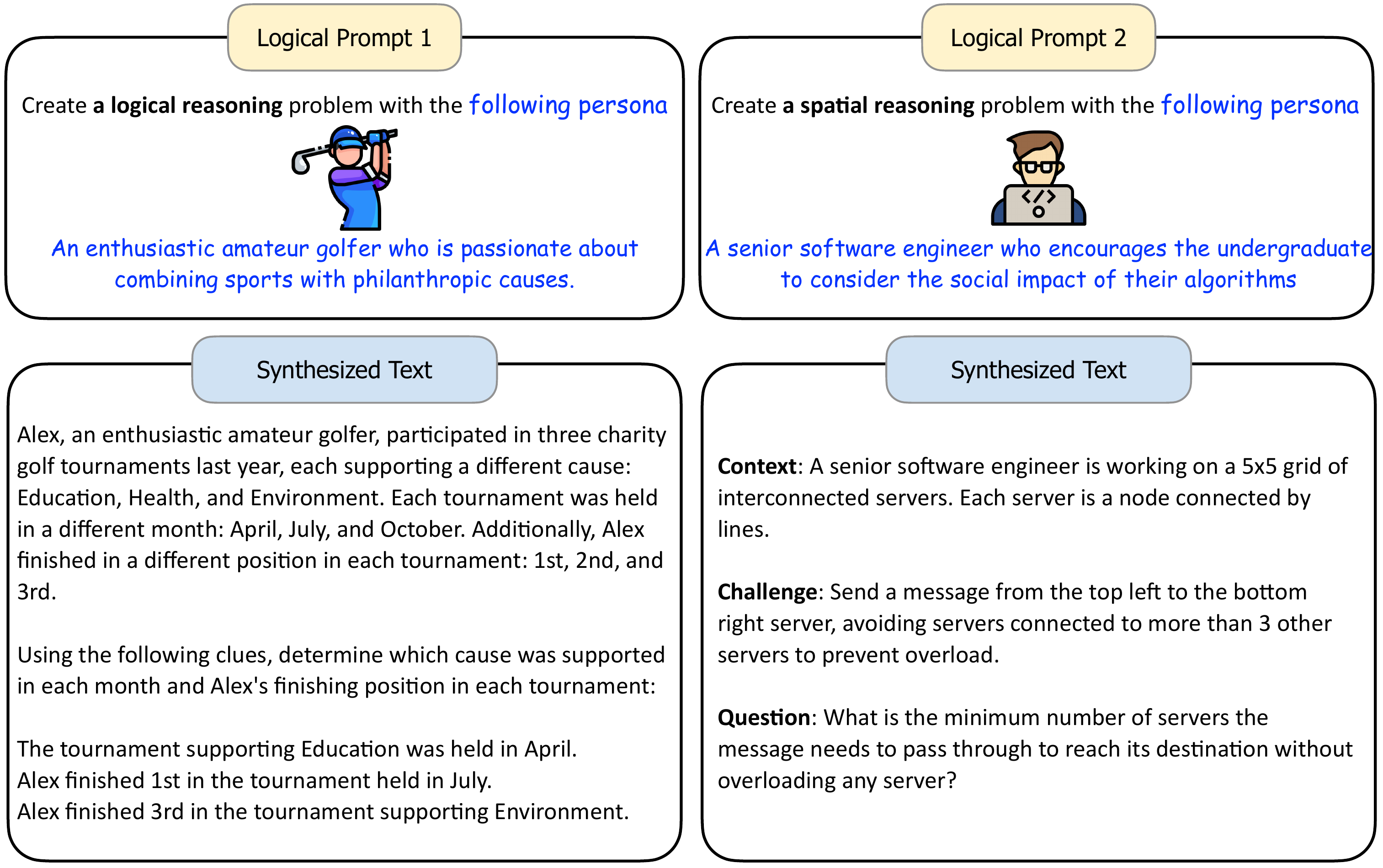}\vspace{-0.2cm}
    \caption{Logical reasoning problems created by our proposed persona-driven methodology}\vspace{-0.2cm}
    \label{fig:logical1}
\end{figure}

\begin{figure}[t]
    \centering
    \includegraphics[width=15cm]{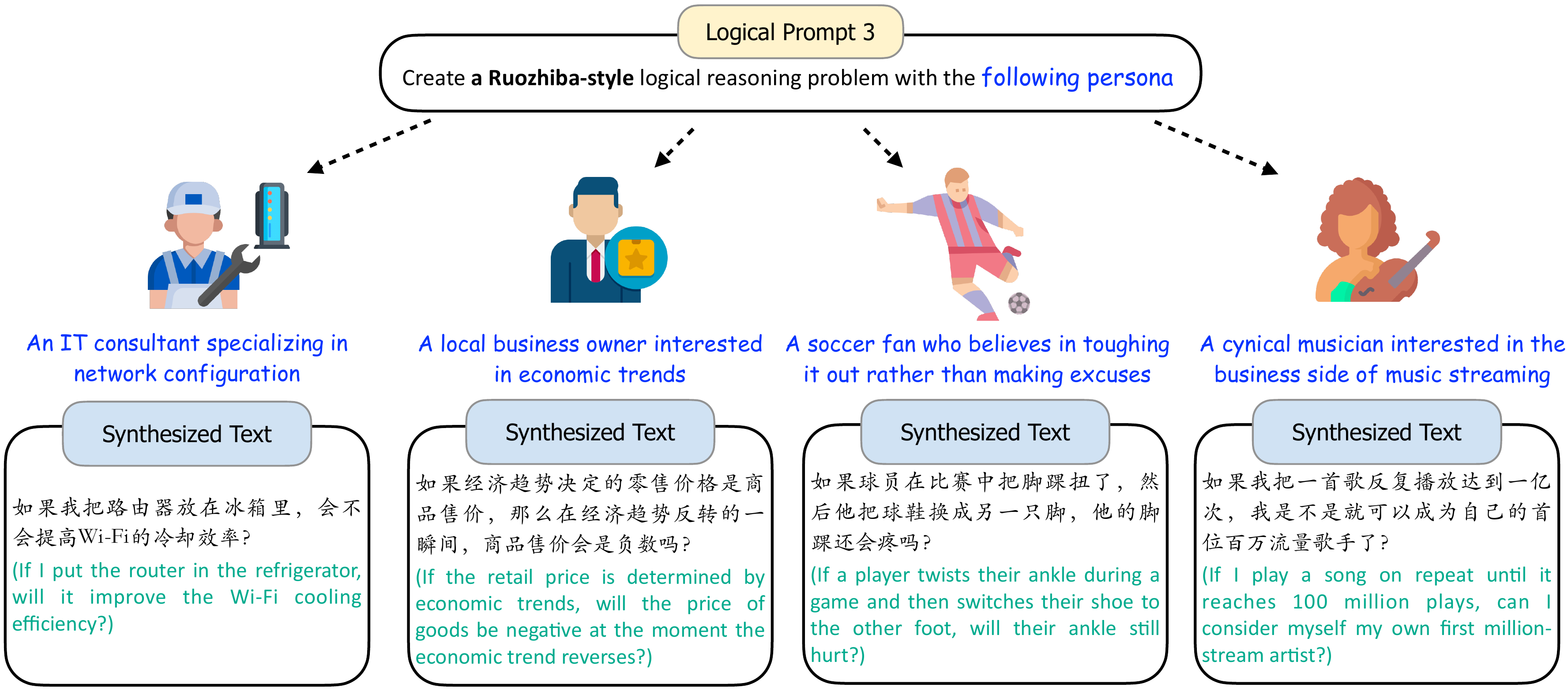}\vspace{-0.2cm}
    \caption{Ruozhiba-style logical reasoning problems created with various personas. Note that Logical Prompt 3 in this figure is a simplified prompt. In practice, we need to specifically define a Ruozhiba-style logical reasoning problem in this prompt in order to obtain desired synthetic data.}
    \label{fig:logical2}
\end{figure}

Similar to math problems, logical reasoning problems can also be easily synthesized. We present examples of typical logical reasoning problems synthesized using our proposed persona-driven methodology in Figure \ref{fig:logical1}. 

\begin{figure}[t]
    \centering
    \includegraphics[width=12cm]{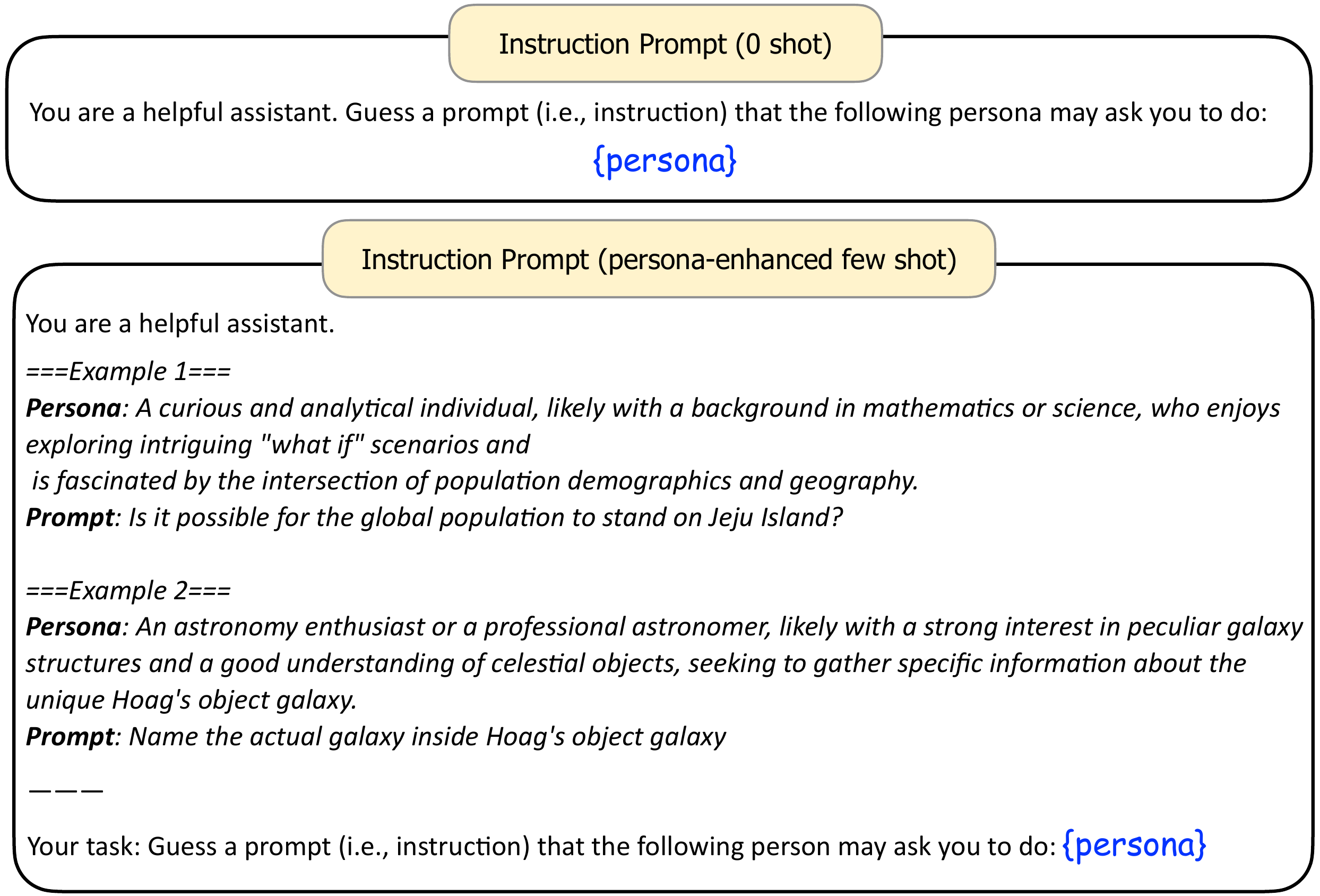}\vspace{-0.2cm}
    \caption{Two typical prompts used for creating instructions (i.e., user prompts).}\vspace{-0.5cm}
    \label{fig:instruction}
\end{figure}

Moreover, we also show several Ruozhiba-style\footnote{\href{https://tieba.baidu.com/f?ie=utf-8&kw=\%E5\%BC\%B1\%E6\%99\%BA\%E5\%90\%A7}{Ruozhiba} is a subforum on \href{https://tieba.baidu.com}{Baidu Tieba}, featuring numerous intricate and challenging questions posted by Chinese netizens. These questions combine elements such as puns, polysemy, causal inversion, and homophones, embedding logical traps that rigorously test the ability to understand complex Chinese language constructs. Recent research \citep{bai2024coig} has demonstrated that these data significantly benefit the improvement of LLMs' logical reasoning abilities.} logical reasoning problems created with personas in Figure \ref{fig:logical2}. All the examples demonstrate that as long as we can clearly describe the requirements for the logical reasoning problem to be created, we can use a large variety of personas to steer the LLM to generate diverse logical reasoning problems that not only meet the requirements but are also highly relevant to the personas, even for whimsical Ruozhiba-style problems.

For more examples, please refer to the 50,000 synthetic reasoning problems we have released.

\subsection{Instructions}\label{subsec:conversation}

The end users of LLMs are ultimately humans. We can use \textrm{Persona Hub} to simulate a variety of users to understand their typical requests for LLM assistance, resulting in diverse instructions (i.e., user prompts).

Figure \ref{fig:instruction} shows two typical persona-driven prompts for synthesizing instructions, corresponding to the zero-shot prompting and persona-enhanced few-shot prompting methods described in Section \ref{sec:methodology}. The zero-shot method does not rely on any existing instruction dataset and allows the LLM to generate various instructions based on different personas. In contrast, the persona-enhanced few-shot method requires existing instruction datasets (e.g., we use WildChat \citep{zhao2024wildchat} in our experiments) to sample some instructions as demonstrations and involves inferring the associated personas of these instructions through the \textit{Text-to-Persona} method described in Section \ref{subsec:t2p}. While this approach is more complex, it results in synthesized instructions that more closely resembles instructions from real users.

With diverse instructions created using \textrm{Persona Hub}, which typically represent the first turn of a user-LLM conversation, we can easily generate their subsequent conversational turns using an LLM, resulting in a large number of simulated user-LLM conversations, which will be valuable for enhancing the LLM's instruction-following and conversational abilities. Furthermore, we can even adopt a similar approach by selecting two personas from \textrm{Persona Hub} and having LLMs role-play both, thereby simulating conversations~\citep{jandaghi2023faithful} between two real people.

As Figure \ref{fig:overview} has already shown some instructions created using this methodology, we skip showing examples here. Readers interested in more examples can refer to the released 50,000 instructions synthesized through 0-shot and persona-enhanced 2-shot prompting.



\subsection{Knowledge-rich Texts} \label{subsec:knowledge}

\begin{figure}[t]
    \centering
    \includegraphics[width=15cm, height=8.8cm]{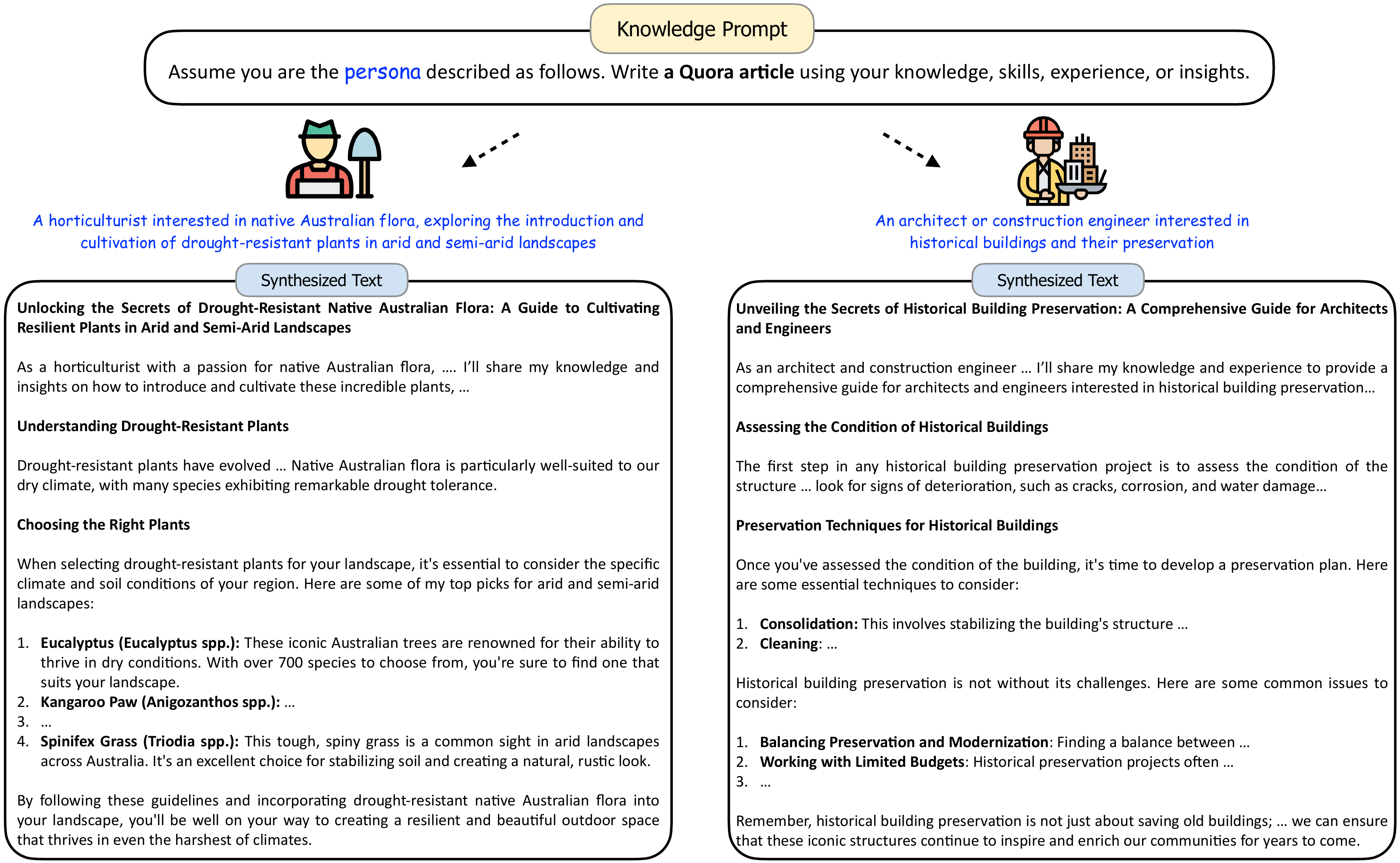}\vspace{-0.2cm}
    \caption{Examples of knowledge-rich plain text synthesis with personas}\vspace{-0.2cm}
    \label{fig:knowledge}
\end{figure}

In addition to synthesizing instructions that can enhance the instruction tuning of LLMs, the persona-driven methodology can be easily adapted to create knowledge-rich plain text that benefits pre-training and post-training of LLMs. As illustrated in Figure \ref{fig:knowledge}, we can prompt an LLM to write a Quora\footnote{\href{https://www.quora.com/}{Quora} is a popular question-and-answer website where users can ask questions and provide answers on a wide range of topics. Articles (i.e., posts) on Quora are often written by knowledgeable individuals, including experts in various fields, ensuring high-quality, well-researched, and informative content.} article\footnote{The methods for synthesizing knowledge-rich plain text are not limited to having the LLM write Quora articles. For instance, we can also prompt the LLM to synthesize (educational) reading material that a persona may be interested in, thereby obtaining a large amount of knowledge-rich text.} using a persona sampled from \textrm{Persona Hub}. This approach elicits the LLM's corresponding knowledge and perspective, resulting in highly informative and knowledge-rich content. By scaling this process with 1 billion personas in \textrm{Persona Hub}, we can easily obtain a vast array of knowledge-rich texts that cover almost any topic across various levels of granularity.

\subsection{Game NPCs} \label{subsec:npc}
A straightforward and practical application of \textrm{Persona Hub} is creating diverse NPCs (Non-Player Characters) at scale for games. As long as we can provide a game's background and world-building information to the LLM, we can prompt the LLM to project personas from \textrm{Persona Hub} (which are typically real-world personas) into characters within the game's world. In this way, we can significantly reduce the effort required for brainstorming NPCs during the game design process.

\begin{figure}[h]
    \centering
    \includegraphics[width=15cm,height=8.5cm]{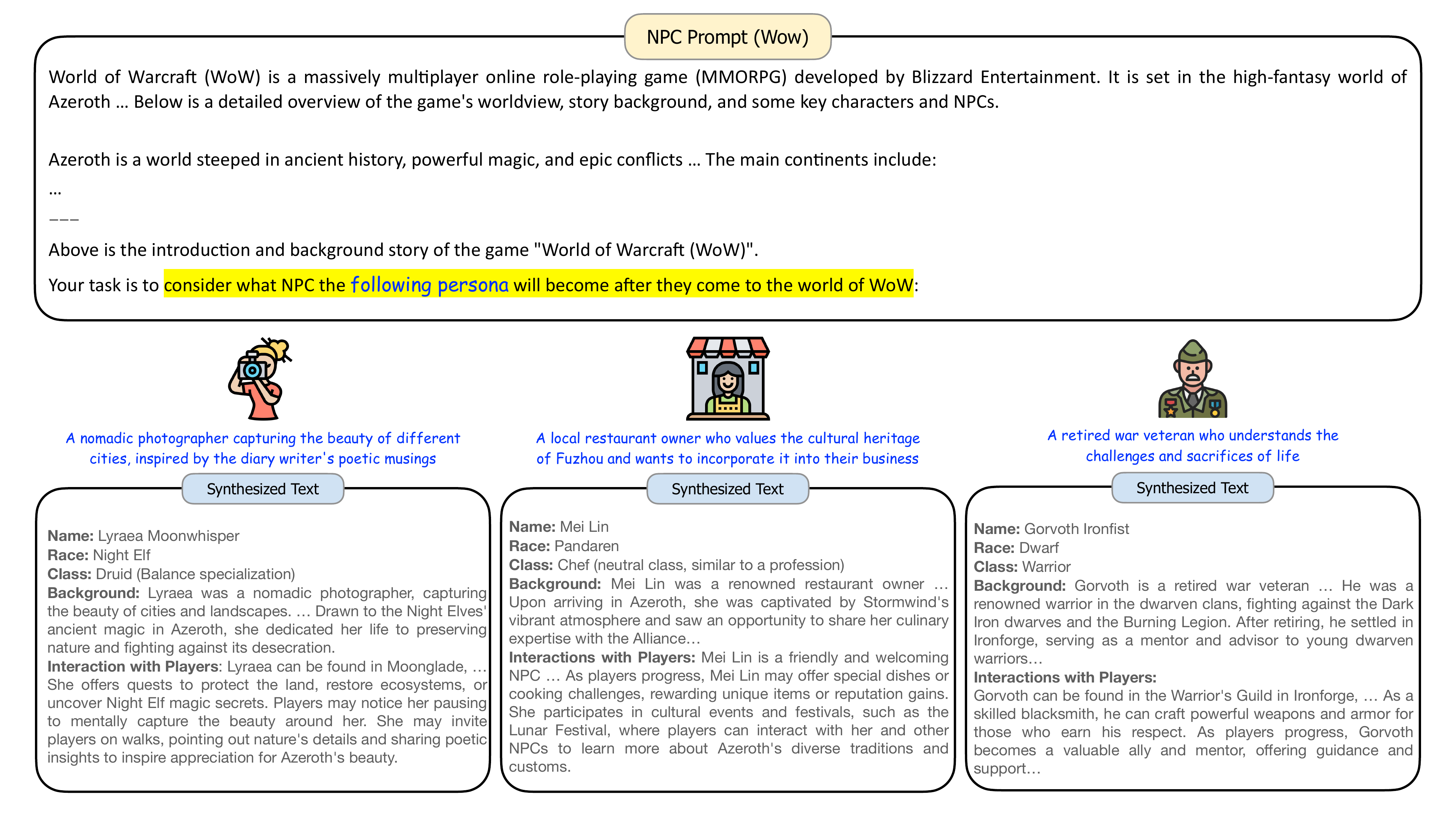}\vspace{-0.2cm}
    \caption{NPC creation for the game ``World of Warcraft'' using personas in \textrm{Persona Hub}}\vspace{-0.4cm}
    \label{fig:wow}
\end{figure}

\begin{CJK}{UTF8}{gkai}

\begin{figure}[t]
    \centering
    \vspace{-0.5cm}
    \includegraphics[width=15cm]{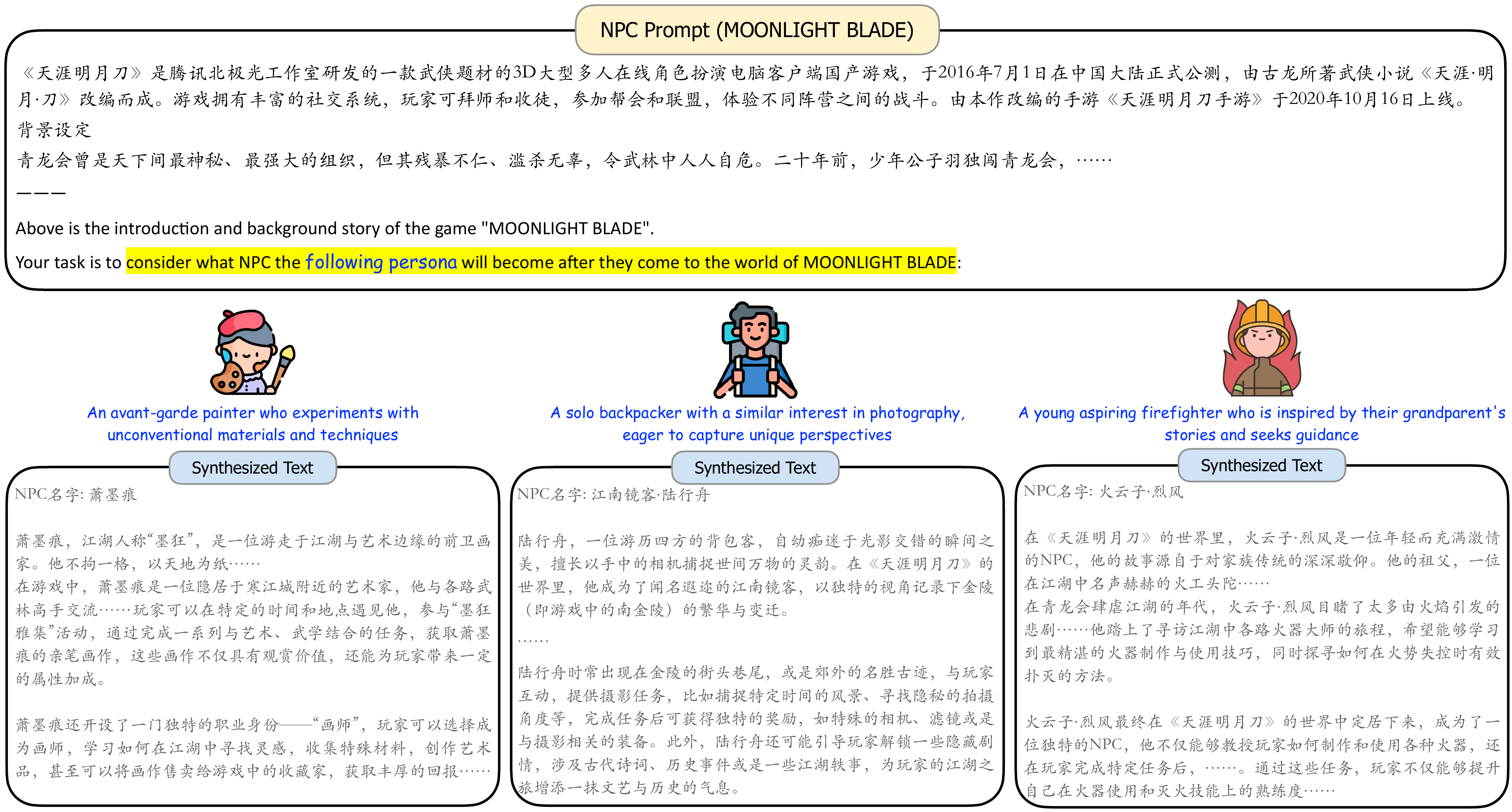}\vspace{-0.2cm}
    \caption{NPC creation for the game ``Moonlight Blade (天涯明月刀)'' using \textrm{Persona Hub}}
    \label{fig:blade}\vspace{-0.2cm}
\end{figure}

Figure \ref{fig:wow} and \ref{fig:blade} show concrete examples where we use personas in \textrm{Persona Hub} to create game NPCs for the game ``World of Warcraft\footnote{\href{https://worldofwarcraft.blizzard.com/en-us/}{World of Warcraft}, developed by Blizzard Entertainment, is a highly influential MMORPG with millions of active players worldwide, spanning over 100 countries since its release in 2004.}'' and ``Moonlight Blade\footnote{\href{https://wuxia.qq.com/main.shtml}{Moonlight Blade (天涯明月刀)} is a 3D martial arts-themed MMORPG developed by Tencent, officially launched in China on July 1, 2016.}''.

\end{CJK}

\subsection{Tool (Function) Development} \label{subsec:tool}

\begin{figure}[htbp]
    \centering
    \includegraphics[width=15cm]{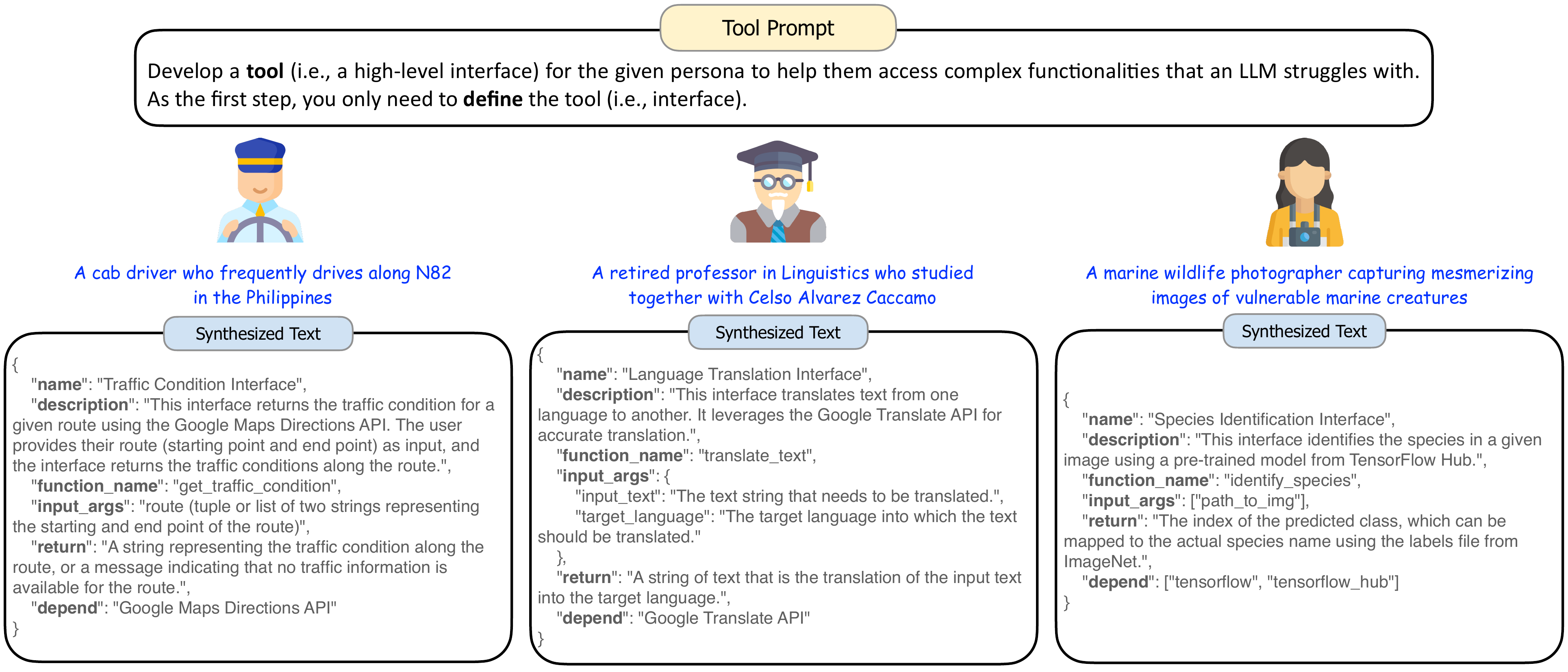}\vspace{-0.2cm}
    \caption{Examples of tool (function) creation with \textrm{Persona Hub}}\vspace{-0.2cm}
    \label{fig:tool}
\end{figure}

As Section \ref{subsec:conversation} demonstrates, \textrm{Persona Hub} can be used to simulate a wide variety of real users to anticipate their possible requests (i.e., instructions) to an LLM. Similarly, we can use \textrm{Persona Hub} to predict the tools \citep{cai2023large,schick2024toolformer} that users might need, so that we can pre-build these tools (functions) beforehand. When a real user makes a similar request, the LLM can directly call these pre-built tools to return results without having to build tools from scratch each time. This paradigm, introduced by Persona Hub, is a completely new solution that allows LLMs to better serve users. We believe it will have great potential in the future as LLMs become more democratized and multifunctional.

\begin{figure}[h]
    \centering
    \includegraphics[width=12cm]{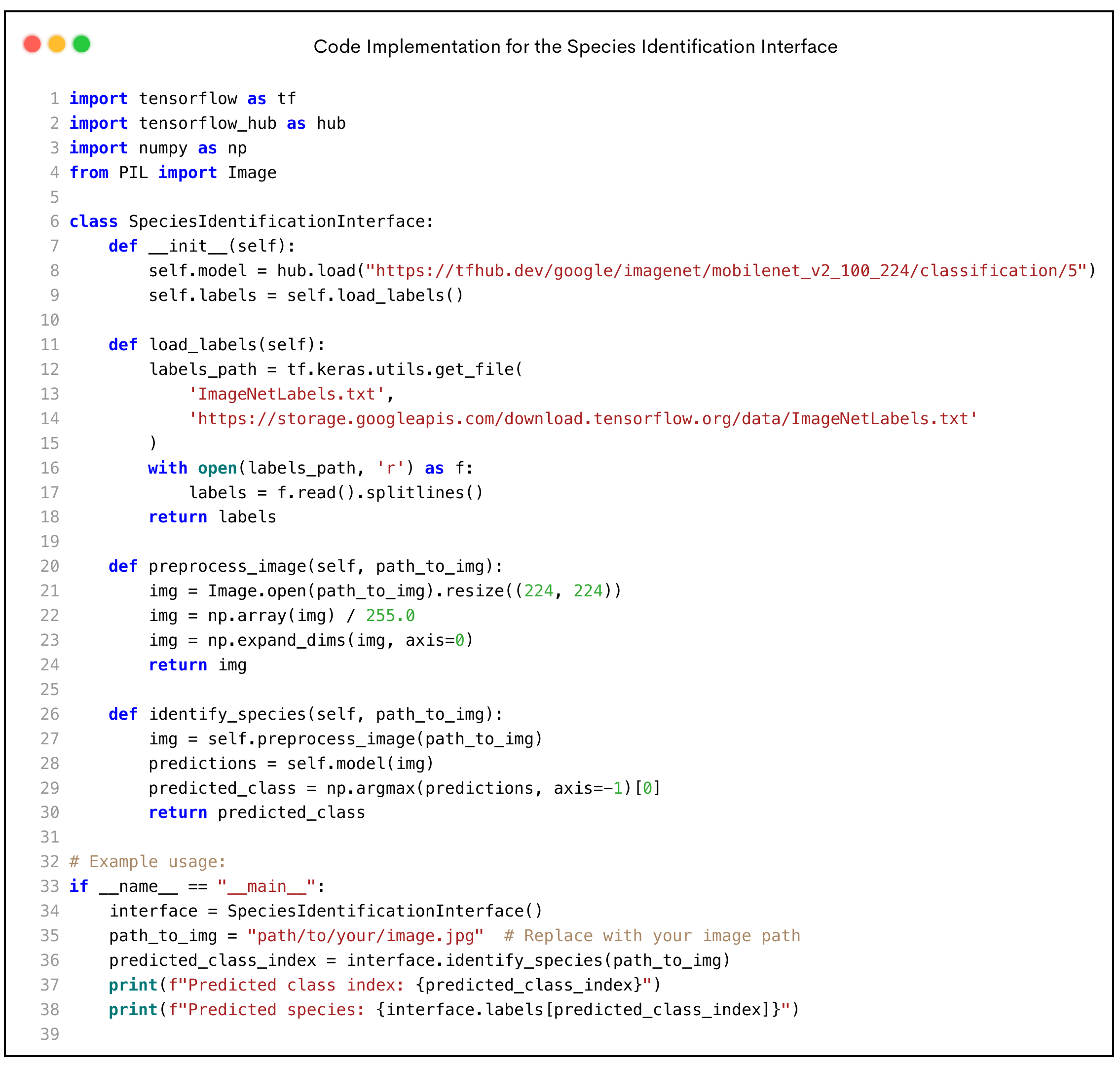}\vspace{-0.1cm}
    \caption{The interface definitions (e.g., the species identification interface) in Figure \ref{fig:tool} can be easily converted into code implementations by calling an LLM to implement them. The resulting pre-built tools can then be directly utilized by the LLM in the future, eliminating the need to build them from scratch each time.} \label{fig:code} \vspace{-0.2cm}
\end{figure}

Figure \ref{fig:tool} shows examples of tools created with various personas. These tools provide functionalities that the personas may need (e.g., a cab driver needs to check traffic conditions) but cannot be accessed by an LLM, greatly expanding the range of services provided by the LLM. Note that although the tools in Figure \ref{fig:tool} are just interface definitions, these definitions can be easily converted into code implementations, as shown in Figure \ref{fig:code}.

\section{Broad Impact and Ethical Concerns}\label{sec:concern}

\subsection{Broad Impact} \label{subsec:impact}

\subsubsection{Paradigm Shift in Data Creation by Humans and LLMs} \label{subsubsec: data_creation}

Traditionally, it has been widely accepted that while LLMs excel at processing data (e.g., rewriting, annotation, or generating outputs/solutions to specific inputs), they are not particularly adept at creating new data. Consequently, the task of data creation has still largely been the domain of humans, and the collaboration paradigm between humans and LLMs has always been humans creating data and LLMs processing it~\citep{maini2024rephrasing}. However, the introduction of our proposed persona-driven methodology potentially revolutionizes this paradigm. With \textrm{Persona Hub}, LLMs are no longer confined to processing existing data; they can now create various types of new data from a multitude of perspectives, much like the diverse population of the world.

While the current capabilities of LLMs may not yet fully replace humans in fulfilling the mission of data creation—whether in terms of data quality or breadth—the ongoing advancements in LLM capabilities suggest a future where LLMs will increasingly excel in data creation. As LLMs continue to improve, both the quality and breadth of the data they can create will also likely enhance, leading us to a point where LLMs may fully take on the role of data creation. When this day arrives, we will no longer be constrained~\citep{villalobosposition} by the limited high-quality human-produced real-world data\footnote{\href{https://lilianweng.github.io/posts/2024-02-05-human-data-quality/}{https://lilianweng.github.io/posts/2024-02-05-human-data-quality/}}. \textrm{Persona Hub} ensures the diversity and coverage of synthetic data, significantly mitigating concerns about the negative impacts~\citep{shumailov2023curse,dohmatob2024tale} of synthetic data on model training. This may effectively eliminate the data bottleneck, thereby pushing the scaling law to its limit.

\subsubsection{Reality Simulation}

In Section \ref{subsec:conversation} and \ref{subsec:tool}, we have demonstrated that \textrm{Persona Hub} can represent a vast array of real-world individuals with its 1 billion personas. By employing these personas to simulate and infer the potential needs and behaviors of real users, we can not only allow LLMs to autonomously prepare for upcoming use cases (queries), but also pave the way for LLMs to effectively mimic the real world, thereby creating many new opportunities. For instance, companies can use this method to predict how different types of users might react to a new product launch; governments can foresee the public's response to new legislation, considering various population group; in online services that require user profiling and behavior modeling, \textrm{Persona Hub} can facilitate the simulation of diverse user behaviors, significantly alleviating the cold start challenge.

Recent research on LLM roleplay, agent collaboration~\citep{liu2023dynamic,wang2024unleashing}, strategic reasoning~\citep{gandhi2023strategic,zhang2024llm}, and related areas can, of course, be facilitated by the vast and diverse personas in \textrm{Persona Hub}. More ambitiously, the 1 billion personas can even sustain a well-organized society within a virtual world, such as sandbox environments~\citep{park2023generative}, online games, parallel worlds, or the metaverse, using the method discussed in Section \ref{subsec:npc} to simulate operations with powerful LLMs. This virtual society can serve as a testing ground for new policies, radical initiatives, and social dynamics, providing valuable insights before real-world implementation. By creating a controlled environment where diverse personas interact, we can observe emergent behaviors, test hypotheses, and refine strategies in a risk-free setting. This can not only help deepen our understanding of complex systems but also speed up innovation by facilitating rapid iteration and experimentation.

\subsubsection{Full Memory Access of LLMs} \label{subsubsec:fullmemory}

When we interact with an LLM in a specific scenario, we can only elicit a fraction of its memory and capabilities, and we are unable to fully access the vast knowledge encapsulated within the LLM. However, \textrm{Persona Hub} potentially offers access to the full memory of an LLM because the 1 billion personas in \textrm{Persona Hub} can tap into almost every perspective and piece of information encoded within the LLM.

By leveraging these 1 billion personas, we can create diverse queries and obtain solutions from a target LLM, thereby transforming the LLM's comprehensive memory (parameters) into synthetic data in textual form. If we consider an LLM as a parameterized compression of world knowledge, then \textrm{Persona Hub} can be viewed as a distributed carrier-based compression\footnote{As Figure \ref{fig:compression} illustrates, Persona Hub has on the order of \(10^{10}\) tokens, equivalent to a $10,000\times$ compression of the public web text (\(10^{14}\) tokens).} of world knowledge, as demonstrated in Figure \ref{fig:compression}. This distributed carrier-based compression provides us with an opportunity to decompress the LLM's parameters back into world knowledge and information it has ever learned (e.g., using the method discussed in Section \ref{subsec:knowledge}).

However, considering that the current \textrm{Persona Hub} is still in a very preliminary stage and that today's LLMs are not yet capable of losslessly converting their memory into synthetic data due to inevitable hallucination \citep{xu2024hallucination}, the breadth and quality of the synthetic data generated through this methodology are still limited. Nevertheless, as \textrm{Persona Hub} continues to improve and scale, and as LLMs become more powerful (with less hallucination), we can look forward to a day when it will be possible to nearly losslessly extract the full memory of an LLM into plain text. 

\subsection{Ethical Concerns}

\subsubsection{Training Data Security and Threats to Current LLM Dominance}

As discussed in Section \ref{subsubsec:fullmemory}, Persona Hub offers an opportunity to access the full memory of a target LLM. However, this also introduces a significant issue: the security of the training data. All data synthesized through the target LLM essentially represents a form of its seen training data. Therefore, the process of extensively extracting a target LLM's memory is essentially dumping its training data, even though this process is generally lossy.

Moreover, if we employ the method described in Section \ref{subsec:conversation} to synthesize instructions (i.e., user prompts) that nearly covers all use cases to query a target LLM to obtain its outputs at scale, there is a high risk that the target LLM's knowledge, intelligence, and capabilities could be extracted and replicated. This poses a challenge to the leading position of the most powerful LLMs, as we have already validated in Section \ref{subsec:math} through mathematical reasoning.

Given that current LLMs generally share similar architectures and their performance advantage primarily lies in their data, this work is likely to impact the current practices and may potentially serve as a turning point, accelerating the shift in the competitive landscape of LLMs from one that heavily depends on data advantage to one that focuses on more advanced technologies.

\subsubsection{Miscellaneous}

Synthetic data presents a general concern of misinformation and fake news, which has been frequently discussed in previous research~\citep{pan2023risk}. Persona Hub potentially amplifies this issue, as diverse personas bring diverse writing styles, making machine-generated texts harder to distinguish from human-generated content~\citep{chakraborty2023possibilities}. This increased difficulty in detection may worsen issues related to data contamination, where synthetic data is mixed with real data, potentially skewing research results and public information.

\section{Conclusion and Future Work} 

We propose a novel persona-driven data synthesis methodology and present \textrm{Persona Hub}, a collection of 1 billion diverse personas automatically curated from web data. We show that this methodology can facilitate the scaling of synthetic data creation across various scenarios, demonstrating its potential to revolutionize creation and applications of synthetic data, and its prospects as a general data synthesis engine for both research and practice.

As the first version of \textrm{Persona Hub}, although it already contains 1 billion personas, the descriptions of these personas are focused only on major aspects and lack fine-grained details (e.g., preferences for colors and numbers; specific family backgrounds, historical contexts, and life experiences). We plan to refine the personas in subsequent versions of \textrm{Persona Hub}, aiming for their descriptions to be as detailed as those found in Wikipedia articles about individuals. These more detailed persona descriptions will make each persona more unique, thereby scaling up Persona Hub and fostering more opportunities for synthetic data creation, while also empowering practical applications such as personalized conversations (e.g., \href{https://character.ai/}{character.ai}).

Also, while this work only explores data synthesis with text-based LLMs, the methodology should also be applicable to multimodal LLMs. Therefore, we will explore multi-modal synthetic data creation as a future direction. Moreover, given that specific personas can elicit corresponding perspectives from LLMs, we are curious about the possibilities of using some super personas to guide LLMs to explore beyond the scope of existing knowledge. This may provide a new approach to tapping into the super intelligence of LLMs, which will be studied in the future.

\bibliography{colm2024_conference}
\bibliographystyle{colm2024_conference}

\end{document}